\pdfoutput=1

\documentclass[11pt]{article}

\usepackage[final]{acl}

\usepackage{times}
\usepackage{latexsym}

\usepackage[T1]{fontenc}

\usepackage[utf8]{inputenc}

\usepackage{microtype}

\usepackage{inconsolata}

\usepackage{float}
\usepackage{framed}
\usepackage{arydshln}
\usepackage{graphicx}
\usepackage{enumitem}
\usepackage{comment}
\usepackage{colortbl}
\usepackage{xcolor}
\usepackage{diagbox}
\usepackage{booktabs}
\usepackage{makecell}
\usepackage{amsmath}
\usepackage{multirow}     
\usepackage{caption}      
\usepackage{array}  
\usepackage{pdfpages}
\usepackage{svg}
\usepackage{longtable}
\usepackage[most]{tcolorbox}
\definecolor{warmnavyblue}{RGB}{25, 61, 89}
\definecolor{con}{rgb}{1, 0.9, 0.9}      
\definecolor{ent}{rgb}{0.9, 1, 0.9}      
\definecolor{neu}{rgb}{0.8, 0.9, 1}      
\definecolor{lightgray}{gray}{0.9}
\tcbset{
  promptstyle/.style={
    enhanced,
    width=\textwidth,          
    colback=gray!10,           
    colframe=black,            
    boxrule=0.5pt,             
    arc=4pt,                   
    left=6pt, right=6pt,       
    top=6pt, bottom=6pt,       
    fontupper=\ttfamily,       
    colbacktitle=black,        
    coltitle=white,            
    title style={font=\bfseries, halign=flush left}, 
    boxed title style={arc=4pt},                   
    attach boxed title to top left={xshift=0mm, yshift=-2mm}, 
  }
}

\newtcolorbox{prompt}[1][]{%
  promptstyle,
  title={Prompt},
  #1
}

\title{FinNLI: Novel Dataset for Multi-Genre Financial Natural Language Inference Benchmarking}

\author{
    \textbf{Jabez Magomere}\textsuperscript{1,2}\thanks{These authors contributed equally to this work.}%
    \thanks{Work done during an internship at JPMorgan AI Research.} 
    \quad
    \textbf{Elena Kochkina}\textsuperscript{2}\footnotemark[1] 
    \quad
    \textbf{Samuel Mensah}\textsuperscript{2} \\
    \textbf{Simerjot Kaur}\textsuperscript{2}
    \quad
    \textbf{Charese H. Smiley}\textsuperscript{2} \\
    \textsuperscript{1}University of Oxford
    \quad 
    \textsuperscript{2}JPMorgan AI Research \\[0.3em]  
    \small{
    \texttt{jabez.magomere@keble.ox.ac.uk} \quad
    \texttt{\{name\}.\{surname\}@jpmorgan.com}
    }
}

\begin{document}
\maketitle

\begin{abstract}

We introduce FinNLI, a benchmark dataset for Financial Natural Language Inference (FinNLI) across diverse financial texts like SEC Filings, Annual Reports, and Earnings Call transcripts. Our dataset framework ensures diverse premise-hypothesis pairs while minimizing spurious correlations. FinNLI comprises 21,304 pairs, including a high-quality test set of 3,304 instances annotated by finance experts. Evaluations show that domain shift significantly degrades general-domain NLI performance. The highest Macro F1 scores for pre-trained (PLMs) and large language models (LLMs) baselines are 74.57\% and 78.62\%, respectively, highlighting the dataset’s difficulty. Surprisingly, instruction-tuned financial LLMs perform poorly, suggesting limited generalizability. FinNLI exposes weaknesses in current LLMs for financial reasoning, indicating room for improvement. 
\end{abstract}

\section{Introduction}

 Large Language Models (LLMs)~\citep{openai2023gpt4,touvron2023llama2openfoundation,touvron2023llama,abdin2024phi3technicalreporthighly}   
excel on a wide variety of tasks.
 However, their generalizability  
 to specialized domains remains an open research question~\citep{li2023chatgpt,zhang2024out}. 
Financial domain presents unique challenges such as financial terminology~\cite{araci2019finbert} and mathematical reasoning~\cite{chen2021finqa}. Thus specialized benchmarks are necessary to ensure models effectively handle the complexities of this field.
Financial Natural Language Processing (NLP) is a growing research area with resources ~\cite{nie2024survey,li2023chatgpt} and models~\cite{yang2023fingpt,yang2023investlm,xie2023pixiu} developed for a wide range of tasks.  However, Natural Language Inference (NLI), one of the core tasks in NLP, has been overlooked in this domain. 

\begin{figure}[t]
\centering
    \includegraphics[width=0.4\textwidth]{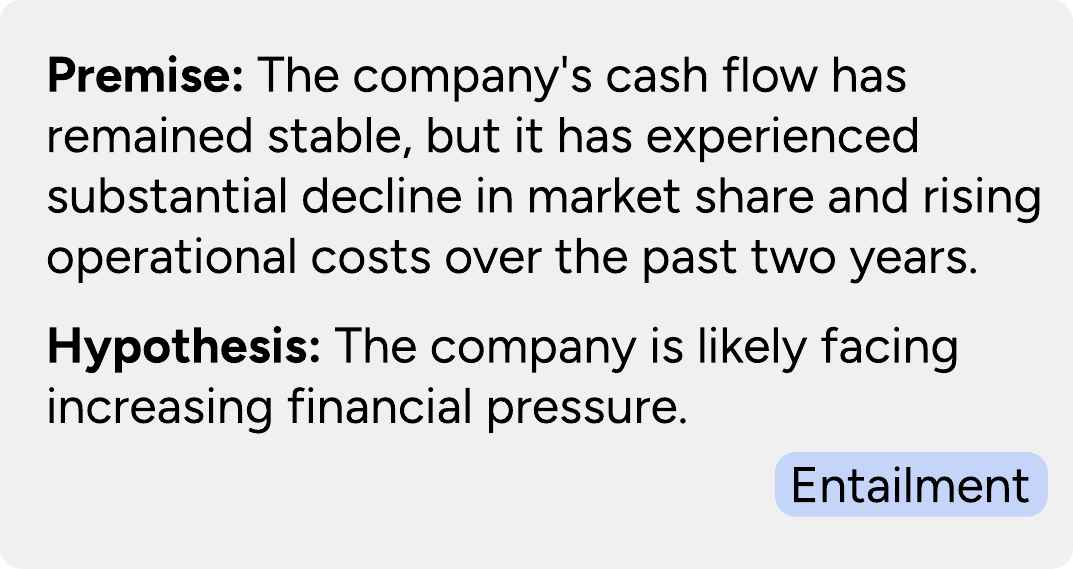}
 \caption{An example of NLI in financial risk assessment, which might seem Neutral as there is no explicit mention of ``financial pressure'' in the premise.} 
 \label{fig:example}
\end{figure}

NLI is a task of determining whether a given hypothesis is true (entailment), false (contradiction), or undetermined (neutral) based on a given premise. 
NLI is essential for downstream tasks such as question answering, fact-checking, stance classification and relation extraction~\citep{samarinas2020latent,chen2021can, sainz-etal-2021-label}.  These are 
widely applicable 
in the financial domain as well, underlying potential applications in risk assessment, stock market prediction, and automated financial reporting.
Figure~\ref{fig:example} shows an example of an NLI task in the financial domain.
While the hypothesis might initially seem neutral due to the lack of explicit mention of ``financial pressure'', a deeper understanding of financial risk assessment shows that stable cash flow, rising operational costs, and declining market share typically signal increasing financial pressure; thus a robust NLI model should classify the pair as entailment.
Without a resource allowing to evaluate the state of the art on financial NLI, the effectiveness of any improvements, or adaptation methods can not be assessed either. 
\begin{figure*}[t]
    \centering
    \includegraphics[width=0.9\textwidth]{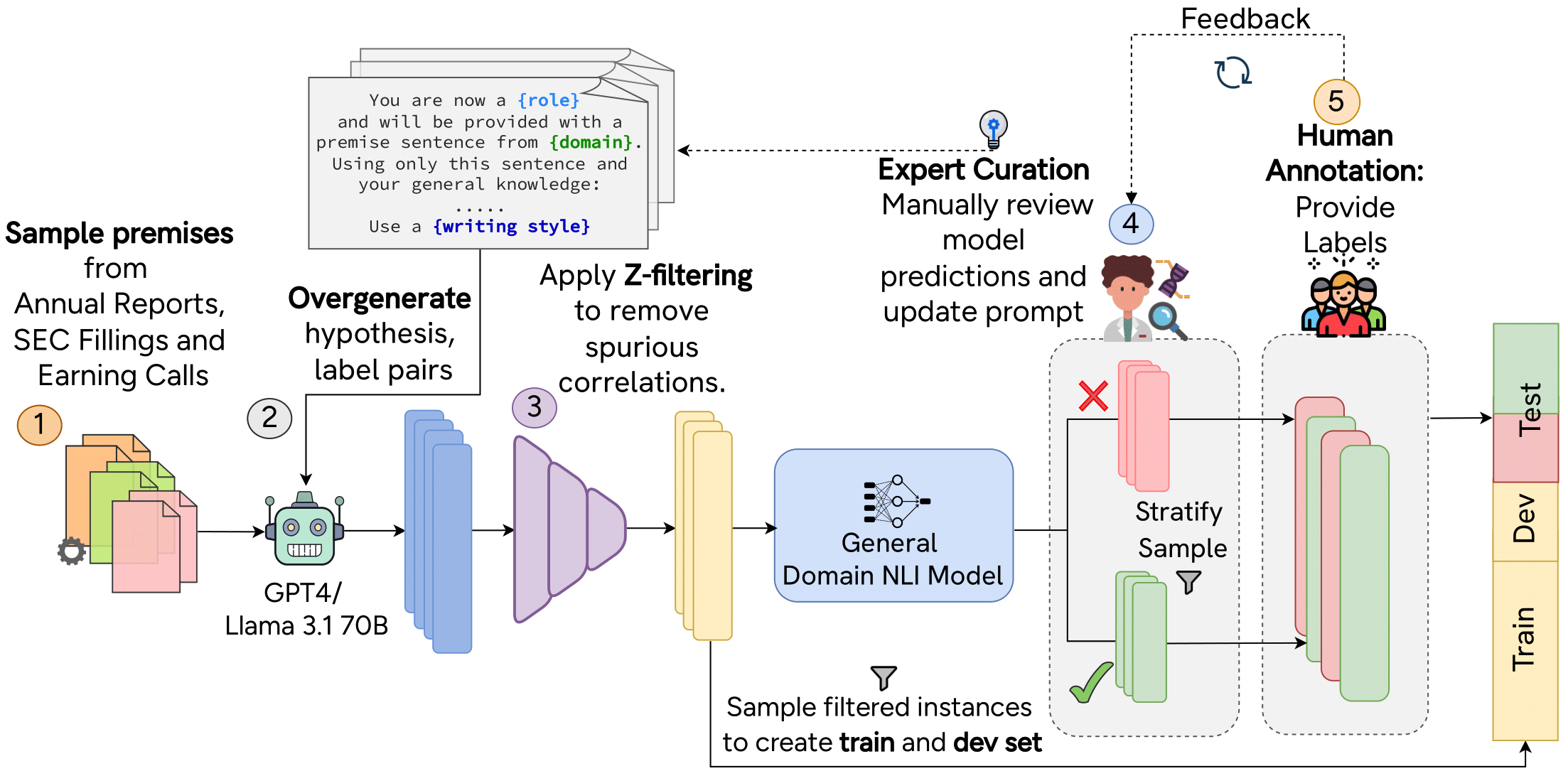}
    \caption{\textbf{Overview of the FinNLI data generation pipeline.}  
    (1) We sample premises from real-world financial documents across multiple genres.  
    (2) Hypothesis-label pairs are generated using multiple LLMs.  
    (3) \textit{Z-filtering}~\cite{wu2022generating} removes spurious correlations.  
    (4) The prompt is refined based on feedback from a general-domain NLI model and expert curation.  
    (5) Finally, instances correctly predicted and misclassified by the NLI model are reviewed by finance experts for gold label annotation.}
    \label{fig:pipeline}
\end{figure*}
To address this gap, we introduce the Multi-Genre \textbf{Fin}ancial \textbf{N}atural \textbf{L}anguage \textbf{I}nference \textbf{(FinNLI)} dataset that enables us to assess and improve the capabilities of language models in understanding and inferring financial information. 

Previous studies have identified artifacts in NLI datasets that can diminish model effectiveness~\citep{gururangan2018annotation,zhang2019mitigating,ye2024spurious}. In this paper, we meticulously design a generation pipeline to address these prevalent challenges.
Our framework for dataset generation shown in Figure~\ref{fig:pipeline} focuses on diversity, mitigation of spurious correlations, including challenging  examples and having high quality annotations for the testing set. 
We then present a comprehensive evaluation of various state-of-the-art models on the FinNLI benchmark. These include general domain and financial LLMs and smaller pretrained language models (PLMs). 
\begin{table*}[t!]
    \centering
    \small
    \resizebox{\linewidth}{!}{
        \begin{tabular}{l|lllllll}
            \toprule
            \textbf{Dataset} & \textbf{Domain} & \textbf{\begin{tabular}[c]{@{}l@{}}Premise \\ Source\end{tabular}} & \textbf{\begin{tabular}[c]{@{}l@{}}Ensures \\ premise \\ diversity\end{tabular}} & \textbf{\begin{tabular}[c]{@{}l@{}}Hypothesis \\ Generation\end{tabular}} & \textbf{\begin{tabular}[c]{@{}l@{}}Ensures \\ hypothesis \\ diversity\end{tabular}} & \textbf{\begin{tabular}[c]{@{}l@{}}Mitigates \\ spurious \\ correlations\end{tabular}} & \textbf{\begin{tabular}[c]{@{}l@{}}Includes \\ challenging \\ instances\end{tabular}} \\ 
            \midrule
            SNLI \cite{bowman2015large} & General & Real-world & No & Human Annotation & No & No & No \\ 
            MNLI \cite{N18-1101} & General & Real-world & Yes & Human Annotation& No & No & No\\ 
            MedNLI \cite{shivade2019mednli} & Medical & Real-world & No & Human Annotation& No & No & No\\ 
            ANLI \cite{nie2019adversarial} & General & Real-world & No & Human Annotation& No & No & Yes\\ 
            WANLI \cite{liu2022wanli} & General & Synthetic & No & Synthetic + Human & No & No & Yes\\ 
            SciNLI \cite{sadat2022scinli} & Scientific & Real-world & No & Rule-Based & No & No & No\\ 
            MSciNLI \cite{sadat-caragea-2024-mscinli} & Scientific & Synthetic & Yes & Rule-Based & No & No & No \\ 
            GNLI \cite{hosseini2024syntheticdataapproachdomain} & General & Synthetic & Yes & Synthetic & Yes & No & No \\ 
            \midrule
            \textbf{Ours (FinNLI)} & \textbf{Financial} & \textbf{Real-world} & \textbf{Yes} & \textbf{Synthetic} & \textbf{Yes} & \textbf{Yes} & \textbf{Yes} \\ 
            \bottomrule
        \end{tabular}
    }
    \caption{Qualitative comparison of NLI dataset generation approaches across different methodological aspects.}
    \label{tab:comparison_with_other_approaches}
\end{table*}
We make the following contributions:
\begin{itemize}
\item We introduce  a novel dataset for evaluating NLI in the financial domain.
\item We present a dataset generation framework and discuss the contribution of its parameters. 
\item We provide a thorough evaluation and its analysis of multiple NLP models on FinNLI.
\item We demonstrate significant room for improvement for Financial NLI  across all models. 
\end{itemize}

\section{Related Work}
\paragraph{Financial NLP}
Financial NLP is a rapidly evolving research area, with a number of advances achieved in a number of tasks such as sentiment analysis~\cite{shah2023trillion,malo2014good}, question answering~\cite{chen2021finqa,chen2022convfinqa}, 
and financial text summarization~\cite{mukherjee2022ectsum}. 
Several finance-specific LLMs have also been developed to address domain-specific challenges. Notable examples include InvestLM~\cite{yang2023investlm}, FinGPT~\cite{yang2023fingpt}, PIXIU FinMA~\cite{xie2023pixiu}, and Bloomberg GPT~\cite{wu2023bloomberggpt},  although access to some of these models is limited, and they are not all available as open-source.While finance-instruction-tuned LLMs have attained increasingly high performance on various financial NLP tasks, it is still unclear whether these models will generalise well to novel tasks not included in the instruction tuning data. The task of financial NLI has been largely overlooked likely due to insufficient resources, yet it has been shown to be informative for evaluating LLMs~\citep{madaan2024lost}. We aim to fill this gap by creating a benchmark dataset and conducting a comprehensive evaluation across a wide range of models.

While there are no dedicated financial NLI datasets, FinCausal datasets~\citep{mariko2020financial, mariko-etal-2022-financial} provide benchmarks for evaluating financial reasoning. However, FinCausal focus solely on causality detection, which is just one type of reasoning. In contrast, NLI task involves various logical relationships that require multiple types of reasoning, with causal reasoning being only one aspect.  
Our proposed dataset, FinNLI, evaluates more types of relationships just beyond causality such as mathematical, temporal and financial.

\paragraph{Domain-specific NLI} 
The development of NLI has been significantly advanced by large-scale datasets such as SNLI~\citep{bowman2015large} and MNLI~\citep{N18-1101}. 
However, a common challenge is the performance drop when transitioning from general-domain models to domain-specific applications~\cite{hupkes2023taxonomy}.   This  highlights the importance of creating and utilizing domain-specific NLI datasets to ensure models are well-adapted to specialized domains.

Specialized NLI datasets have been developed for various domains, including biomedical (BioNLI~\citep{bastan2022bionli}), medical (MedNLI~\citep{shivade2019mednli}), scientific (SciNLI~\citep{sadat2022scinli}, SciTail~\citep{scitail}), and legal (ContractNLI~\citep{koreeda2021contractnli}, LegalNLI~\citep{yang2022legalnli}) contexts. 
In this work, we introduce a Financial NLI dataset to evaluate and enhance NLI model performance in the financial domain, where precise inferential reasoning is crucial. 
\paragraph{NLI Dataset Generation}
Approaches to NLI dataset construction typically involve either creating both premise and hypothesis~\cite{hosseini2024syntheticdataapproachdomain}, sampling existing premises and generating hypotheses~\cite{N18-1101,shivade2019mednli}, or sampling both from existing texts~\cite{sadat2022scinli,bastan2022bionli}. The generation process can be automated or manual, often followed by human annotation~\cite{liu2022wanli}. Sampling both premises and hypotheses from existing texts can be limited by rule-based methods (e.g., using linking phrases~\cite{sadat2022scinli,sadat-caragea-2024-mscinli}), as suitable pairs are not easy to find and automatically sample. Human-generated hypotheses are more costly due to the time required. Recent approaches use LLMs to synthetically generate both premises and hypotheses ~\cite{liu2022wanli, hosseini2024syntheticdataapproachdomain}. However, while this may be effective for general domains, LLMs trained on general text may fail to generate high-quality data that accurately capture the nuances of specialized domains~\cite{xu2023knowledgeinfusedpromptingassessingadvancing}. To address these limitations, our work samples premise sentences from real-world documents across multiple financial genres to ensure premise diversity and to induce finance-specific knowledge. We ensure hypothesis diversity by employing different writing styles and roles when prompting LLMs~\cite{yu2023large}, followed by domain-expert annotation to create a high-quality test set.
Table \ref{tab:comparison_with_other_approaches} provides a qualitative comparison of our NLI dataset generation approach with others, further comparisons with finance and NLI datasets are in Appendix~\ref{app:comparison}.

NLI datasets often contain artifacts—spurious correlations between labels and features that do not represent the underlying task \cite{stacey-etal-2020-avoiding, cosma-etal-2024-hard}. These artifacts cause models to learn shortcuts, failing to generalize to new data~\cite{ye2024spurious}. Artifacts can occur in both manually and automatically generated instances~\cite{gururangan2018annotation,zhang2019mitigating,arakelyan2024semantic, proebsting2024hypothesisonlybiaseslargelanguage}. Addressing artifacts involves careful dataset design, filtering~\cite{wu2022generating}, and adding adversarial samples~\cite{nie2019adversarial,liu2022wanli}. In contrast to other approaches, we incorporate z-filtering~\cite{wu2022generating} in our dataset generation pipeline to mitigate spurious correlations and leverage feedback from a general-domain NLI model to ensure our resulting dataset includes challenging instances in an adversarial-like setting

\section{Dataset Construction}
We present our data generation framework (Figure~\ref{fig:pipeline}) which is applicable across various domains. This framework ensures contextually grounded instances by sampling premises from real-world financial texts. It promotes diversity by incorporating two different LLMs and varied prompts while minimizing artifacts through \textit{z-filtering}~\cite{wu2022generating}. To enhance the dataset's difficulty, we leverage feedback from a general domain NLI model and experts to guide model generation. Finally, we maintain high quality by involving multiple finance experts in the annotation process.

\subsection{Stage 1: Sampling Premises Across Multiple Financial Domains}
As the sources of our premise sentences, we use publicly available financial documents: \textbf{Annual Reports}~\footnote{https://www.annualreports.com/}, \textbf{SEC Filings}~\footnote{https://www.sec.gov/} 
and \textbf{Earning Calls}
~\cite{CIKM2020MAEC}. Sampling from diverse financial genres enhances the stylistic variation of the dataset. 
Annual Reports and SEC Filings exhibit a formal style with prevalence of numerical information~\cite{loukas-etal-2022-finer}, while Earning Calls  include more colloquial language.  
Preprocessing is used to filter out invalid sentences, ensuring sample quality (Appendix~\ref{app:preproc}).

\subsection{Stage 2: Generating Hypotheses}\label{sec:generating hypotheses}
Given a premise, we prompt an LLM to generate one entailment, neutral, and contradiction hypothesis. The prompt includes the NLI task definition, label verbalization, domain-specific generation conditions, and exemplars to guide the LLM. Conditions and examples are refined through iterations of steps 1-4 of the pipeline on Figure~\ref{fig:pipeline} (full prompt in Appendix~\ref{app:prompt}).
To encourage stylistic variations, we use conditional prompting with role-playing~\cite{shanahan2023roleplaylargelanguagemodels} and treat writing style as an attribute dimension~\cite{yu2023large}. We define a list of finance-related roles as \textit{financial analyst}, \textit{financial reporter}, \textit{finance compliance officer}, and \textit{financial consultant}, and a list of writing styles as \textit{social media}, \textit{news}, \textit{financial textbook}, and \textit{financial reporting}. At each generation step, a role and writing style are randomly selected to condition the generation. We experiment with two LLMs to generate hypothesis-label pairs: GPT-4~\cite{openai2023gpt4} and Llama 3.1 70B~\cite{dubey2024llama3herdmodels}. These models belong to different LLM families, vary in size, and perform competitively across a wide range of tasks. This setup allows us to examine potential differences between closed-source and open-source models in generating synthetic instances.

\subsection{Stage 3: Automatic Filtering}\label{sec:automatic_filtering}
To minimise spurious correlations that could have appeared during the generation process, 
we apply the \textit{z-filtering} algorithm from~\citet{wu2022generating}. It is a method to reject samples that contribute to the high spurious correlations between task-independent features of the samples and their labels. 
The list of task-independent features includes: unigrams and bigrams, lexical overlap between the premise and hypothesis, hypothesis length and premise-hypothesis length ratio, and the number and density of financial terms in the hypothesis to capture correlations that may result from the presence of specific financial keywords. We set the parameters of the z-filtering process, the top \(k\) features and sample batch size to \(20\) and \(200\) respectively. We observed that a higher value of \(k\) and lower value of batch size lead to a higher number of samples to be removed and lower z-scores. Z-scores and top-k features before and after applying z-filtering are included in Appendix \ref{app:z_filtering} showing the effectiveness of this our approach in minimizing spurious correlations. 
We create the train and development sets by randomly sampling from the filtered instances generated by Llama 3.1 70B with the final optimised prompt\footnote{We use Llama 3.1 70B model to generate the training and development set as its license permits the use of the generated artefacts for fine-tuning downstream models \url{https://llama.meta.com/llama3/license/}}. 
We sample evenly across financial domains and class labels. We use the generated labels as the gold standard for training and development. 

\subsection{Stage 4: Expert Curation}\label{sec:expert_curation}
Evaluating the quality of open-ended data generation across various prompt parameters is a challenging task due to the large number of possible prompt combinations \cite{long2024llmsdrivensyntheticdatageneration,chung-etal-2023-increasing}. To address this, we combine feedback from a general domain NLI model and human experts to identify label inconsistencies and challenging reasoning patterns. These insights are then used to iteratively refine the generation prompt. 
As a general domain NLI model we chose RoBERTa-Large NLI\footnote{https://huggingface.co/joeddav/xlm-roberta-large-xnli}, trained on MNLI and SNLI as one of top performing models on this task. Through multiple iterations, we refine the prompt to include instructions for mathematical/quantitative reasoning, temporal reasoning, financial knowledge reasoning, and tricky linguistic constructs, along with representative examples from each reasoning pattern. We provide examples of instances with challenging reasoning patterns in Appendix \ref{tab:challenging_examples}.

We create a domain- and class-balanced test set by sampling instances where the predicted label matches the generated label, as well as instances where they differ. This ensures a mix of correct and incorrect predictions, balancing both challenging and core cases. By doing so, we capture not only ``hard'' examples but also fundamental NLI reasoning patterns, allowing for a more effective evaluation of the generation model~\cite{bowman2021fixbenchmarkingnaturallanguage}.

\subsection{Stage 5: Human Annotation}
We provide the samples for annotation in \(4\) rounds (R) and at each round collect feedback from the expert annotators on the quality of the instances. We provide annotators with instances generated from GPT-4 in R1 and R2 while Llama generated instances in R3 and R4. We incorporate this feedback by updating the prompt conditions and in-context examples to improve the quality of subsequent generations after each annotation round. The annotators are provided with a premise and hypothesis sentence and asked to provide an appropriate label, a confidence level (\textit{high} or \textit{low}), an optional flag to mark the premise or hypothesis sentence as incomplete/nonsensical and an optional comment field. The annotators do not revise the premise or hypothesis sentences  (Appendix~\ref{app:annotation}). 
Each premise-hypothesis pair is reviewed by 3 professional annotators with finance backgrounds and we assign the majority label as the gold label. In very rare cases \((0.15\%)\) where all the annotators assign different labels, we do not assign a gold label. We exclude any instances where \textit{any} of the annotators mark the premise or hypothesis as incomplete/nonsensical.

\begin{table}[t]
\centering
\resizebox{0.8\columnwidth}{!}{%
    \begin{tabular}{@{}ccc@{}}
        \toprule
        \textbf{Split} & \textbf{Size} & \textbf{Label Distribution (E/N/C)} \\ \midrule
        Train          & 16,200        & 5,400 / 5,400 / 5,400                \\
        Dev            & 1,800         & 600 / 600 / 600                     \\
        Test           & 3, 304         & 1,361/ 995 / 948\\ \bottomrule
    \end{tabular}
    }
    \vspace{-2mm}
    \caption{Dataset Split and Label Distribution}
    \vspace{-4.5mm}
    \label{table:data_distribution}
\end{table}

\section{Dataset Analysis}
We provide an analysis of the annotation outcomes and the resulting characteristics of FinNLI\footnote{For information on the data license and dataset access, please contact the authors via email.}.

\subsection{Synthetic Label Accuracy}
\label{sec:agreement_rate}
\textbf{Per Annotation Round.}
We conduct annotations over four rounds, altering either the prompt or the model between rounds. Rounds R1 and R2 utilize GPT-4, while R3 and R4 use hypotheses generated by Llama 3.1 70B. Prompt adjustments occur between R1 and R2, as well as between R3 and R4. In the first round of annotation (R1), domain experts evaluated \(1,462\) GPT-4-generated instances.  In \(65.18\%\) of the cases, the label assigned by the annotators matched the generated label. After refining the prompt, 
the second round (R2) showed improvement, with \(76.47\%\) out of \(595\) instances displaying label agreement, highlighting the value of incorporating expert feedback. 
In annotation rounds three (R3) and four (R4), annotators were presented with examples generated by Llama 3.1 70B. In R3, \(64.57\%\) of the \(621\) instances  had labels that matched the generated ones. 
The decrease in agreement relative to R2 can be attributed to the use of a GPT-4-optimized prompt with Llama, revealing model-specific prompt sensitivity \cite{lu-etal-2024-prompts}. 

After refining the prompt for Llama in R4, label agreement improved to \(76.84\%\) across \(626\) instances. We use the prompt from R4 to generate the training and development sets.  \(14.27\%\) of the examples where the general-domain NLI model assigns a similar label to the generated label are relabelled by annotators, compared to \(46.88\%\) for instances with differing labels. This finding supports the rationale to sample from both correctly and incorrectly predicted instances, as the former likely includes `easy' examples and the latter contains both `hard' and `ambiguous' cases.

The overall agreement rate for the FinNLI dataset is noticeably lower than the \(84.97\%\) \footnote{We obtain the overall agreement rate by averaging the agreement rate for \textit{unanimous} labels - \(89.53\%\) and majority labels - \(80.41\%\) reported in \citet{hosseini2024syntheticdataapproachdomain}} agreement reported for general-domain synthetic NLI datasets such as GNLI \cite{hosseini2024syntheticdataapproachdomain}. Although, we note that \citet{hosseini2024syntheticdataapproachdomain} prompt-tune a generator LLM with an existing NLI dataset while our approach relies solely on expert-curated prompt conditions and in-context exemplars to steer the generator. We also hypothesize that LLMs may be more efficient data generators for general domains than in specialised settings \cite{xu2023knowledgeinfusedpromptingassessingadvancing}. 

\textbf{Per class.}
The breakdown by class label shows significant differences. 
For contradiction hypotheses, both LLMs achieve high agreement rates (GPT-4 - \(88.83\%\) and Llama - \(90.86\%\)). For entailment hypotheses, Llama achieves a higher agreement rate of \(75.46\%\) compared to \(53.55\%\) for GPT-4. In contrast, for neutral hypotheses, GPT-4 achieves a higher agreement rate of \(81.07\%\) compared to \(66.52\%\) for Llama. Qualitative observations show that many LLM-generated entailment hypotheses are relabelled to neutral due to models appending additional information not present in the premise, resulting in lower agreement rates for entailment.

\label{sec:dataset_statistics}
\subsection{Dataset Statistics}
The domain-experts annotated a total of \(3,360\) examples, out of which \(3,304\) instances \((98\%)\) are marked as valid examples. For the valid instances, we achieved a Fleiss-\(\kappa\) score of \(88.31\%\), indicating an almost perfect agreement. 
Table \ref{table:data_distribution} outlines the size of the train, development and testing sets, as well as label distribution across the dataset splits. The training and development sets are uniformly balanced across labels, whereas the test set shows a slight skew toward the entailment label. 
The slight imbalance in the test set is a result of reassigning labels as part of the manual annotation process. 

\label{sec:linguistic_analysis}
\subsection{Linguistic Analysis}
We provide a detailed linguistic analysis of the FinNLI dataset in Table \ref{table:linguistic_analysis} and comparison to other NLI datasets in Appendix \ref{app:comparison}. 
\begin{table}[t]
\resizebox{\columnwidth}{!}{%
\begin{tabular}{@{}lccccccl}
\toprule
       & \multicolumn{2}{c}{\textbf{\#Words}} & \multicolumn{2}{c}{\textbf{\% Sentences}} & \textbf{Word}  &\\
                \cmidrule(lr){2-3} \cmidrule(lr){4-5}
               \textbf{Domain}      & \textbf{Prem.} & \textbf{Hyp.}  & \textbf{Prem.} & \textbf{Hyp.}  & \textbf{Overlap}  & \textbf{Agrmt.}\\ \midrule
\textsc{Reports} & 25.0&  20.6&  95.2  & 98.8 & 41.62\%  &76.14\%\\
\textsc{SEC}   & 23.2 & 20.3& 96.8  & 98.9   & 45.26\%  &76.20\%\\
\textsc{Calls}  & 19.5 & 17.5& 96.0  & 99.2 & 27.92\%   &77.62\%\\ \midrule
\textbf{Overall} & \textbf{22.6}& \textbf{19.5}& \textbf{96.0} & \textbf{98.9} & \textbf{38.25\%}  & \textbf{76.65\%}\\ \bottomrule
\end{tabular}
}
\caption{Dataset Statistics for FinNLI by genre. \textit{\#Words} is the mean number of tokens in the premise and hypothesis.\textit{`\%-Sentences'} -- proportion of sentences parsed with an `S' root using the Stanford PCFG Parser \cite{klein-manning-2003-accurate}. \textit{Word Overlap}  -- mean proportion of tokens shared between the premise and hypothesis. \textit{Agrmt.} is the \% of examples where the generated label matches the label assigned by annotators.}
\label{table:linguistic_analysis}
\end{table}
Premise sentences are generally longer (mean \(22.6\) words, max \(105\) words) than hypothesis sentences (mean \(19.5\) words, max \(84\) words). Premises from Annual Reports and SEC Filings tend to be lengthier, reflecting the formal nature of financial reporting, whereas premises from Earning Calls are shorter, indicative of more colloquial language use. Annual Reports and SEC Filings often include tables and forms, which can result in incomplete premise sentences when sampled. However, our pre-processing pipeline filters out these invalid sentences, with approximately $\approx 96\%$ of the premise sentences containing `S'-root parses, indicating they are syntactically complete. We observe a higher token overlap between the hypothesis and premise sentences in Annual Reports and SEC Filings compared to Earning Calls. Annual Reports and SEC Filings often include specific financial jargon that must be reflected in the hypothesis to remain relevant, whereas utterances in Earning Calls are more open to interpretation.

\section{Models}
\begin{table*}[h] 
\centering
\scalebox{0.7}{
  \begin{tabular}{l c c c c c c }
    \toprule
{\textbf{Model}} & {\textbf{Size}} & {\textbf{Setup}} &  {\textbf{Annual Reports}} &  {\textbf{SEC Filings}} & {\textbf{Earning Calls}} & {\textbf{Overall}}\\
    \midrule
    \textsc{DeBERTav3-NLI}&  \(304M\) & \textsc{Zero-Shot} & $68.83 \pm 0.0$ & $68.52 \pm 0.0 $ & $71.08 \pm 0.0 $ & $69.54 \pm 0.0$  \\
    \textsc{BART-Large-MNLI}&  \(406M\)  & \textsc{Zero-Shot}& $69.96 \pm 0.0$ & $70.24 \pm 0.0$ & $\bf75.95 \pm 0.0$ & $72.09 \pm 0.0$ \\
    \midrule
    \textsc{RoBERTa-base} &  \(255M\) &  \textsc{Finetuned} & $70.32 \pm 0.1$ & $69.14 \pm 1.2$ & $69.84 \pm 0.2 $ & $69.78 \pm 0.5$ \\
    \textsc{FiLM} & \(255M\) & \textsc{Finetuned} & $70.43 \pm 0.6$ & $69.57 \pm 0.2$ & $70.24 \pm 0.8$ & $70.11 \pm 0.01$\\
    \textsc{RoBERTa-large}& \(355M\) & \textsc{Finetuned} & $\bf 75.11 \pm 0.1$ & $73.98 \pm 1.2$ & $74.52 \pm 0.2$ & $\bf74.57 \pm 0.5$ \\
    \midrule
    \textsc{BART-large-MNLI}& \(406M\) & \textsc{Finetuned} & $ 73.91 \pm 0.8$ & $\bf74.14 \pm 0.6$ & $74.35 \pm 0.9$ & $74.18 \pm 0.2$ \\
    \bottomrule
  \end{tabular}}
\vspace{-2mm}
   \caption{Average Macro F1 Scores (\%) of zero-shot general-domain NLI models and fine-tuned PLMs on the FinNLI test set across each financial domain. Results are averaged over three random seeds, with the standard deviation reported. Best performance is highlighted in \textbf{bold}.}
\label{table:baseline_results}
\end{table*}
We evaluate performance of various models on FinNLI test set. We use the average macro F1 as our main metric due to the test set class imbalance. 

\subsection{Zero-Shot NLI Baselines}
We evaluate two state-of-the-art general-domain NLI models: \textbf{BART-Large-MNLI} \cite{lewis-etal-2020-bart}, trained on MNLI~\cite{N18-1101} and \textbf{DeBERTa-V3-NLI} \cite{he2021debertav3}, trained on both MNLI and SNLI \cite{bowman2015large}.

\subsection{PLM Baselines}
We fine-tune the following general-purpose and domain-specific pre-trained language models (PLMs) on the FinNLI training set: \textbf{RoBERTa-Base}, \textbf{RoBERTa-Large} \cite{DBLP:journals/corr/abs-1907-11692}, and \textbf{FiLM} \cite{choe-etal-2023-exploring}, which is a RoBERTa-Base model further pre-trained on financial corpora. Additionally, we fine-tune \textbf{BART-Large-MNLI} \cite{lewis-etal-2020-bart} to explore whether training on both MNLI and FinNLI yields any benefits. 
We conduct a grid search for hyperparameters and choose the best model on the development set. The best model is trained over three independent runs with different random seeds
\footnote{Full details on the experimental setup and hyperparameter tuning are provided in Appendix~\ref{app:params}.}. 

\subsection{LLM Baselines}
We investigate the performance of general-purpose and domain-specific LLMs across a wide range of model sizes and prompt settings on the FinNLI test set. For the general-purpose LLMs, we evaluate  \textbf{Llama-2-7b-chat-hf }\cite{touvron2023llama2openfoundation}, \textbf{Phi-3.5-mini-instruct }\cite{abdin2024phi3technicalreporthighly}, \textbf{Llama-3.1-8B-Instruct} \cite{dubey2024llama3herdmodels}, and \textbf{Llama-3.1-70B-Instruct} \cite{dubey2024llama3herdmodels}. For the finance-domain LLMs, we evaluate \textbf{FinMA-7B-NLP}, \textbf{FinMA-30B} \cite{xie2023pixiu} - Llama-2 7B and 30B respectively instruction-tuned on financial NLP tasks. 
FinMA models were chosen due to superior performance across diverse financial NLP tasks~\cite{xie2024finben} and its open-source availability. 
We use several prompt setups (Appendix~\ref{app:prompt}): Zero-Shot (ZS); Zero-Shot expanded with Annotation Guidelines (ZS-AG); Few-Shot with Annotation Guidelines (FS-AG) (\(2\) examples per class from different financial genres); and Chain-of-Thought (CoT) \cite{wei2023chainofthoughtpromptingelicitsreasoning}. 

\section{Results}\label{sec:results}
\subsection{Zero-shot and Fine-tuned PLM Baselines}
Table~\ref{table:baseline_results} presents the results of experiments with zero-shot NLI and fine-tuned PLM Baselines.

\textbf{General-domain NLI models exhibit a sizeable performance drop on financial NLI.}
BART-Large-MNLI and DeBERTa-V3-NLI achieve accuracy scores of \(89.9\%\) \cite{lewis-etal-2020-bart} and \(91.1\%\)\footnote{Accuracy score reported from \url{https://huggingface.co/cross-encoder/nli-deberta-v3-large}}, respectively, on MNLI. However, their accuracy drops significantly to \(70.94\%\) and \(68.31\%\) on FinNLI, highlighting the impact of domain shift on performance in out-of-domain settings. The general-domain NLI models perform better on Earnings Calls compared to Annual Reports and SEC Filings, likely due to the MNLI training data, which includes telephone transcripts and speeches, aligning more closely with the informal language used in Earnings Calls~\cite{N18-1101}.

\textbf{Fine-tuning PLMs on FinNLI improves performance over zero-shot NLI models.}
Despite the FinNLI test set consisting of GPT-4 and Llama3.1-generated instances while the training set consisting of only Llama3.1-generated instances, we observe performance improvements post fine-tuning. Fine-tuning improves performance by \(5\%\) in F1 score when comparing the zero-shot DeBERTa-V3-NLI to the fine-tuned RoBERTa-Large on the FinNLI test set, both having similar parameters. BART-Large-MNLI shows a \(2\%\) improvement in performance after fine-tuning on FinNLI. However, combining MNLI and FinNLI offers no benefits compared to training solely on FinNLI. Larger fine-tuned PLMs outperform smaller models, suggesting that FinNLI contains challenging examples that require greater model capacity for accurate inference. Domain-specific pre-training offers only marginal gains over general-purpose PLMs. FiLM \cite{choe-etal-2023-exploring}, a RoBERTa-Base model pre-trained on financial documents,\footnote{Including \(301M\) tokens from SEC Fillings and \(1.1B\) tokens from Earning Calls \cite{choe-etal-2023-exploring}} performs slightly better than the standard RoBERTa-Base.

\begin{figure*}[h!]
    \centering
    \includegraphics[width=0.9\textwidth]{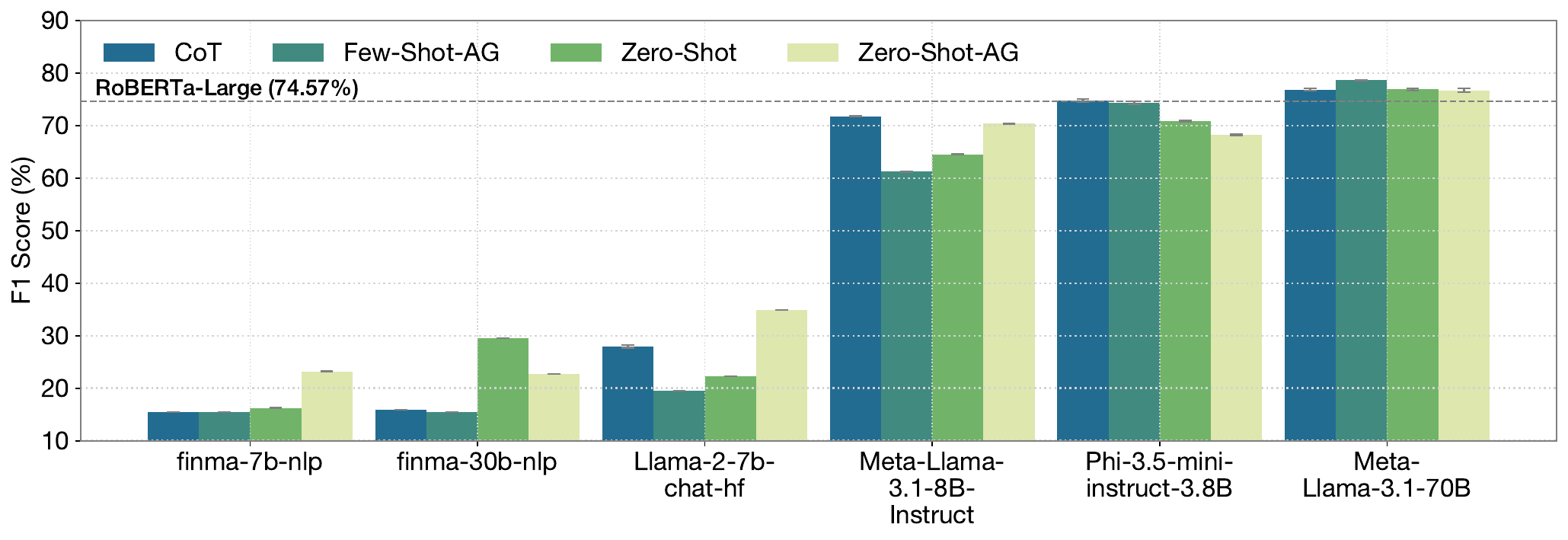}
    \vspace{-4mm}
    \caption{Average macro F1 scores (\%) for various LLMs across different prompting setups evaluated on the FinNLI test set. Prompts with \textbf{AG}  have the class label definitions used in the annotation guidelines included in the prompt. The error bars represent the standard deviation across \(3\) independent runs. The dotted grey line at \(74.57\%\) marks the performance of RoBERTa-Large, the best performing fine-tuned PLM on FinNLI test set. }
    \label{fig:llmresults}
\end{figure*}

\subsection{LLM baseline results}
Figure~\ref{fig:llmresults} shows the results of LLM experiments. 

\textbf{Llama 3.1 70B achieves the best performance on FinNLI, but significant room for improvement remains.}
Llama 3.1 70B achieves the highest F1 score of \(78.62\%\) on the FinNLI test set and is the only LLM among those evaluated that outperforms a finetuned RoBERTa-Large model, which attains an F1 score of \(74.57\%\) (\(\sigma = 0.5\%\)). Phi 3.5 achieves a comparable performance of \(74.74\%\) (\(\sigma = 0.3\%\)), with this difference being too small to conclude any significant advantage over RoBERTa-Large. All other LLMs are outperformed by the RoBERTa-Large model finetuned on the FinNLI training set, indicating that PLMs are still highly competitive on domain-specific tasks. 
While the best performing F1 score is \(78.62\%\), we observe that this is notably lower than other LLMs' performance on general-domain NLI tasks \cite{wang-etal-2024-rethinking} \footnote{General-domain NLI datasets are likely to be included in LLM pretraining data and as such this may not be a fair evaluation} suggesting that FinNLI may contain examples that exhibit complex reasoning patterns or contain `ambiguous' instances that pose a challenge to current state-of-the-art LLMs. To investigate this, we conduct an error analysis on the instances misclassified by  Llama 3.1 70B in Section \ref{sec:error_analysis}.

\textbf{Small Language Models (SLMs) outperform significantly larger LLMs on FinNLI.} 
Phi-3.5-mini-instruct, a 3.8 billion parameter LLM, achieves an F1 score of \(74.74\%\), outperforming Llama 3.1 8B, a significantly larger LLM that achieves an F1 score of \(71.79\%\). This underscores the potential of smaller language models trained on highly curated datasets in bridging the performance gap to larger LLMs \cite{abdin2024phi3technicalreporthighly}.

\textbf{LLMs instruction-tuned on financial domain tasks perform worse than general-domain LLMs.}
FinMA 7B and FinMA 30B - instruction-tuned on financial NLP tasks, achieve F1 scores of 23.20\% and 29.46\% respectively. In contrast, the base model for FinMA 7B, Llama-2-7B-chat achieves a higher F1 score of 34.87\%. 
This suggests that instruction-tuning may have negatively impacted the model's ability to generalise to other tasks outside the instruction-tuning data. 
This finding aligns with previous studies which show that domain-specific instruction tuning can degrade general performance on financial textual-analysis tasks \cite{xie2024finben} and knowledge-based evaluation benchmarks \cite{lee_finale_2024}. Possible reasons for this could include limited training sets, suboptimal training parameters, or training schedules~\citep{dong2023abilities,lee2024instruction}.

\textbf{CoT prompting improves performance for small to mid-sized LLMs on FinNLI.}
We observe that using CoT and asking an LLM to provide an explanation for its prediction, leads to performance improvements over zero-shot prompting for small to mid-sized models on FinNLI. Specifically, CoT  yields an approximate F1Score improvement of \(7\%\) for Llama 3.1 8B, \(5.6\%\) for Llama-2-7b-chat-hf, and \(3.9\%\) for Phi-3.5-mini-instruct. However, for the larger Llama 3.1 70B model, CoT does not provide significant performance improvement over zero-shot. 
Score). 
The FinMA models do not exhibit improvements with CoT as they fail to return any meaningful responses. We observe that smaller LLMs are more sensitive to different prompts on the NLI task, and CoT helps enhance their reasoning. 
In contrast, larger LLMs like Llama 3.1 70B already demonstrate strong reasoning capabilities and are less affected by prompt variations.

\section{Impact of Generation Design Decisions}
We discuss
the impact of 
different generation design decisions on model performance. 

\subsection{GPT-4 vs Llama-Generated Test Sets}
We compare the average macro F1 scores of the best-performing zero-shot NLI model, fine-tuned PLM and LLM on the GPT-4-generated instances \((2,057)\) and Llama-generated instances \((1,247)\), averaged over three independent runs.
BART-Large-MNLI performs significantly better on the Llama test set (\(80.43\%\)) than on the GPT-4 test set (\(65.19\%\)) suggesting that GPT-4-generated instances are more challenging for zero-shot NLI models. Fine-tuned RoBERTa-Large achieves higher performance on the GPT-4 test set (\(77.48\%\)) compared to the Llama test set (\(71.90\%\)). Fine-tuning appears to help the model adapt to the nuances of GPT-4-generated data but reduces its effectiveness on Llama-generated instances.
Llama 3.1 70B exhibits consistent performance across both test sets (\(77.28\%\) on GPT-4 and \(78.84\%\) on Llama)\footnote{A paired t-test confirms that the higher performance on the Llama test set is statistically significant (\(t = 50.59\), \(p = 0.0004\)).}. The slight advantage  on the Llama test set could be attributed to self-preference bias, where an LLM's performance is favourable to its own output as compared to other outputs \cite{panickssery2024llmevaluatorsrecognizefavor}.

\subsection{Prompt Roles and Writing Styles}
We investigate how different prompt roles and writing styles used to generate hypotheses affect model performance. For the best-performing fine-tuned PLM (RoBERTa-Large) and LLM (Llama 3.1 70B), we observe no significant differences in F1 scores across the various roles or styles, indicating robustness to stylistic variations. However, the zero-shot NLI model, BART-Large-MNLI, shows considerable performance discrepancies. It achieves higher F1 scores on more formal writing styles such as \textit{financial textbook} (\(78.32\%\)) and \textit{financial reporting} (\(77.40\%\)) compared to less formal styles like \textit{social media} (\(60.70\%\)) and \textit{financial news} (\(73.82\%\)). This suggests that zero-shot NLI models are more sensitive to stylistic variations \cite{belinkov-etal-2019-dont}. In contrast, LLMs like Llama 3.1 70B gain robustness to different writing styles due to pretraining on diverse domains \cite{dubey2024llama3herdmodels}. Similarly, the fine-tuned PLM gains robustness by training on the FinNLI, enabling better generalization across stylistic variations.

\subsection{High vs Low Confidence Annotations}
We define high-confidence instances as examples that receive a unanimous label and all annotators mark their confidence level as high for that instance. Low-confidence instances are examples where at least one annotator marks with low confidence or instances that do not have a unanimous label. 
Out of \(3,304\) annotated examples, \(2,401\) examples (\(72.67\%\)) are classified as high-confidence and \(903\) examples (\(27.33\%\)) marked as low-confidence. 
Llama 3.1 70B achieves an F1 score of \(82.51\%\) on the high-confidence instances compared to \(67.20\%\) on the low-confidence instances. This  suggests that LLMs similarly struggle on examples that human annotators demonstrate low confidence or high disagreement, possibly the ambiguous cases.

\label{sec:error_analysis}
\section{Error Analysis}
Reviewing the frequency of model's errors across the classes, we observe a notable trend of misclassifying neutral instances as entailment (Appendix ~\ref{app:further_error_analysis}).
To gain a deeper understanding, we manually reviewed \( \approx10\%\) (\(60\) out of \(683\)) of the misclassified instances from the best-performing model, Llama 3.1 70B(Appendix~\ref{app:challenging_instances}). Our analysis shows that \(56\%\) of misclassifications are due to \textbf{inference errors} (clear model mistakes), while \(44\%\) stem from \textbf{ambiguous instances}, i.e., phrases or words that allow more than one interpretation~\cite{sandri-etal-2023-dont,jiang-marneffe-2022-investigating}. For inference errors, the model often assumes \textit{implicit information} (\(26\%\)) not contained in the premise, making logical inferences that, while plausible, are unfaithful to the premise~\cite{zhou-etal-2023-context}. Additionally, the model struggles with instances requiring \textit{financial reasoning} (\(16\%\)), such as knowledge of financial terms or standard procedures, which domain-expert annotators easily pick up. \textit{Tricky linguistic instances} \cite{nie2019adversarial} (\(8\%\)) such as counter-speech~\cite{gligoric-etal-2024-nlp}, words play, and those requiring \textit{mathematical reasoning} (\(6\%\)) still pose a challenge to the model. Regarding ambiguous instances, \(22\%\) of misclassifications are due to \textit{uncertainty in sentence meaning} that may result in multiple interpretations. Interestingly, a significant proportion (\(22\%\)) of premise sentences sampled from \textit{Earnings Calls} contain \textit{anaphora} and \textit{deixis} \cite{poesio-artstein-2005-reliability}, typical of informal conversations —expressions that require contextual information to interpret. Such instances lead to ambiguity, as it is often unclear whether the entity introduced in the hypothesis is mentioned in the premise, resulting in \textit{missing context}.

\section{Conclusion}
We introduced the FinNLI benchmark to evaluate and improve NLI models in the financial domain. Our evaluations show that general-domain NLI models perform worse on financial NLI, LLMs perform comparably to fine-tuned PLMs, and instruction-tuned financial LLMs lack generalizability. This reveals significant room for improvement in financial NLI, necessary for better understanding and reasoning with financial text, ultimately enhancing downstream applications.

\section{Limitations}
Our NLI task setup employs a single gold label classification, and we have yet to explore more advanced setups, such as those involving explanation rationales or generating probability distributions over the label space that align with human judgment. Due to limited domain expertise, we could only employ three annotators per instance.

We did not explore adversarial settings, which could provide insights into model robustness against attacks. Future work will explore domain adaptation methods and using FinNLI as an intermediate task to enhance performance on related tasks. Currently, the FinNLI benchmark is limited to English, restricting its applicability to non-English financial texts. We also note that the generated premises may contain fictional information and should not be used to train models that learn factual data.

\section{Ethical Considerations}
The financial documents utilized in the FinNLI dataset, including SEC filings, annual reports, and earnings call transcripts, are publicly available and do not contain sensitive, personally identifiable information (PII). Earnings call transcripts have been anonymized. However, it is important to acknowledge that financial texts and model-generated outputs may contain inherent biases. Despite our efforts, we cannot guarantee the complete absence of biases and hallucinations in the synthetic data produced. These limitations should be considered when interpreting the results and applications of our research. This dataset should not be used directly for tasks like fact-checking and question answering, as we cannot guarantee the accuracy of the information. 

\section*{Acknowledgments}
We would like to thank Arturo Oncevay, Toyin Aguda, and Keshav Ramani for insightful discussions. We thank our colleagues from Annotation Center of Excellence, who collaborated with us to refine annotation guidelines and produce high quality annotations for this research paper.

\paragraph{Disclaimer}This paper was prepared for informational purposes by the Artificial Intelligence Research group of JPMorgan Chase \& Co. and its affiliates ``JP Morgan'') and is not a product of the Research Department of JP Morgan. JP Morgan makes no representation and warranty whatsoever and disclaims all liability, for the completeness, accuracy or reliability of the information contained herein. This document is not intended as investment research or investment advice, or a recommendation, offer or solicitation for the purchase or sale of any security, financial instrument, financial product or service, or to be used in any way for evaluating the merits of participating in any transaction, and shall not constitute a solicitation under any jurisdiction or to any person, if such solicitation under such jurisdiction or to such person would be unlawful. 
© 2025 JPMorgan Chase \& Co. All rights reserved 

\bibliography{bibliography}

\appendix
\label{sec:appendix}
\section{Comparison to other Datasets}
\label{app:comparison}
\begin{table*}[t]
\centering
\small
 \resizebox{\linewidth}{!}{
  \begin{tabular}{ l r r r r r r r r r }
    \toprule
      &  & \multicolumn{3}{c}{\bf \#Examples} & \multicolumn{2}{c}{\bf \#Words} & \multicolumn{2}{c}{\bf \% Sentences} & \textbf{Word} \\
       \cmidrule(lr){3-5}  \cmidrule(lr){6-7}  \cmidrule(lr){8-9}
   {\bf Dataset }  & {\bf Domain } & {\bf Train}     & {\bf Dev}   & {\bf Test} & {\bf Prem.} & {\bf Hyp.} & {\bf Prem.} & {\bf Hyp.} & {\bf Overlap}  \\ 
   \midrule
   SNLI \cite{bowman2015large} & General &550,152 & 10,000 & 10,000 & 14.1 & 8.3 & 74.0\% & 88.9\% & 52.97\% \\  
   MNLI \cite{N18-1101} & General & 392,702 & 20,000 & 20,000 & 22.3 & -\% & 91.0\% & 98.0\% & - \\
   ANLI \cite{nie2019adversarial} & General & 162,865 &  3,200 &  3,200 & -\% & -\% & -\% & -\% & -\% \\
   WANLI \cite{liu2022wanli} & General & 102,885 & - & 5,000 & -\% & -\% & -\% & -\% & -\% \\
   GNLI \cite{hosseini2024syntheticdataapproachdomain} & General & 670,739 & 6,845 & 490 & 40.5 & 11.0 & -\% & -\% & -\% \\
   MedNLI \cite{shivade2019mednli} & Medical & 11,232 &  1,395 & 1,422 & 20.0 & 5.8 & -\% & -\% & -\% \\
   SciNLI \cite{sadat2022scinli} & Scientific & 101,412 & 2,000 & 4,000 && 25.93 & 96.8\% & 96.7\% & 30.06\%\\
   MSciNLI \cite{sadat-caragea-2024-mscinli} & Scientific & 127,320 & 1,000 & 4,000 & 26.84 & 25.85 & 94.4\% & 94.3\% & 30.29\% \\ 
\midrule
\textbf{FinNLI (Ours)}  & Financial & 16,200 & 1,800 & 3,304 & 22.6 & 19.5 & 96.0\% & 98.9\% & 38.25\%\\
    \bottomrule
  \end{tabular}
  }
  \caption{Comparison of key statistics of FinNLI with other related datasets. 
  \textit{\#Example} is the number of instances across each data split - we report the human-annotated test size for equal comparison. \textit{\#Words} is the mean number of tokens in the premise and hypothesis.\textit{`\%-Sentences'} -- proportion of sentences parsed with an `S' root using the Stanford PCFG Parser \cite{klein-manning-2003-accurate}. \textit{Word Overlap}  -- mean proportion of tokens shared between the premise and hypothesis.}
    \label{table:dataset_statistics_comparison}
\end{table*}
We provide a detailed comparison of FinNLI to other NLI datasets in Table \ref{table:dataset_statistics_comparison} and other Financial NLP datasets in Table \ref{tab:comparison_with_other_datasets}. FinNLI is the first NLI dataset specifically focused on the financial domain, filling a significant gap in existing datasets and providing a benchmark for financial reasoning tasks. When comparing FinNLI's size with other NLI datasets, our evaluation set's size is comparable to other domain-specific NLI datasets (Table \ref{table:dataset_statistics_comparison}) and larger than most existing financial NLP datasets (Table \ref{tab:comparison_with_other_datasets}). General-domain NLI datasets tend to have larger test sizes, while  smaller test sizes are observed in domain-specific datasets reflecting the time and costs incurred in obtaining high-quality expert annotations. Although FinNLI's training dataset is smaller compared to other datasets, the training and development sets are sufficient for model learning on the task. This is evidenced by model improvements observed when fine-tuned on the FinNLI training set (see results in Section \ref{sec:results}). Additionally, our data generation approach is scalable and can generate more instances with additional resources.

FinNLI's premise and hypothesis pairs are generally longer than those in general-domain datasets, capturing the unique linguistic complexity of financial texts such as Annual Reports and SEC Filings. This length difference may contribute to the performance drop observed in models trained on general-domain NLI datasets when applied to FinNLI (shown in Section \ref{sec:results}). The lexical overlap between premise and hypothesis sentences in FinNLI is lower than in general-domain datasets like SNLI and comparable to other domain-specific NLI datasets. This lower overlap indicates that our dataset contains fewer spurious features which models could otherwise exploit to excel on the task.
\begin{table}[ht]
\centering
\resizebox{\linewidth}{!}{
\begin{tabular}{l l r}
    \toprule
    \textbf{Dataset}                      & \textbf{Task}                             & \textbf{Test Size} \\
    \midrule
    FPB~\citep{malo2014good}               & Sentiment Analysis                        & 970       \\
    FiQA-SA~\citep{maia2018www}           & Sentiment Analysis                        & 235       \\
    TSA~\citep{cortis2017semeval}            & Sentiment Analysis                        & 561       \\
    Headlines~\citep{sinha2021impact}  & News Headline Classification              & 2,283     \\
    FinCausal~\citep{mariko2020financial}      & Binary Classification                      & 8,630     \\
    FOMC~\citep{shah2023trillion}            & Hawkish-Dovish Classification             & 496       \\
    FinArg-ACC~\citep{sy2023fine}        & Argument Unit Classification              & 969       \\
    FinArg-ARC~\citep{sy2023fine}     & Argument Relation Classification          & 496       \\
    MultiFin~\citep{jorgensen2023multifin} & Multi-Class Classification                & 690       \\
    MA~\citep{yang2020generating}              & Deal Completeness Classification          & 500       \\
    MLESG~\citep{chen2023multi}            & ESG Issue Identification                  & 300       \\ \midrule
    \textbf{FinNLI (ours)}                & \textbf{Natural Language Inference}       & \textbf{3,304}       \\
    \bottomrule
  \end{tabular}
}
\caption{Comparison of FinNLI's evaluation set with other available Financial NLP textual analysis datasets.}
\label{tab:comparison_with_other_datasets}
\end{table}

\subsection{Model Performance Comparison to other Datasets}
We compare the performance of RoBERTa-Base and RoBERTa-Large on FinNLI and other NLI datasets, as shown in Table \ref{tab:roberta_performance_comparison}. Across the NLI datasets compared, the performance of both RoBERTa-Large and RoBERTa-Base is lower on FinNLI. This suggests that our dataset is potentially ``more difficult'' than other NLI datasets, likely due to the specialized language, longer premise-hypothesis pairs and unique reasoning required in the financial domain. However, we note that realistically comparing the difficulty of different NLI datasets can be tricky. This is because various factors, such as how the training data is generated and annotated, the size of the datasets, the distribution of labels, and the evaluation metrics used, can all vary.  
\begin{table}[H]
\centering
\resizebox{\linewidth}{!}{
\begin{tabular}{l l r l}
    \toprule
    \textbf{Dataset}     & \textbf{Domain}        & \textbf{Accuracy (\%)} & \textbf{Source}                             \\
    \midrule
    \textsc{\textbf{RoBERTa-Large}} \\
    \midrule
    SNLI             & General           & 91.40                      & \cite{sun2020selfexplainingstructuresimprovenlp}           \\
    MNLI             & General           & 90.20                      & \cite{zhuang-etal-2021-robustly}                           \\
    MedNLI           & Medical           & 83.30                      & \cite{lewis-etal-2020-pretrained}                          \\
    WANLI            & General           & 75.40                      & \cite{liu2022wanliworkeraicollaboration}                   \\
    \textbf{FinNLI (Ours)} & \textbf{Financial}     & \textbf{75.37}           &                                                  \\
    \midrule
    \textsc{\textbf{RoBERTa-Base}} \\
    \midrule
    SciNLI          & Scientific        & 78.12                      & \cite{sadat2022scinli} \\  
    MSCINLI          & Scientific        & 77.42                      & \cite{sadat-caragea-2024-mscinli}                         \\
    \textbf{FinNLI (Ours)} & \textbf{Financial}     & \textbf{70.69}           &                                                  \\
    \bottomrule
  \end{tabular}
}
\caption{Comparison of dataset performance across different domains using RoBERTa models.}
\label{tab:roberta_performance_comparison}
\end{table}

\section{Premise Preprocessing} 
\label{app:preproc}
We observe that directly sampling premise sentences from financial documents without applying any preprocessing results in a high number of invalid sentences. These include: (1) incomplete fragments, possibly caused by parsing errors from the original documents, (2) sentences containing tables, (3) sentences dominated by numerical data, (4) titles and subtitles, and (5) complete but irrelevant sentences, such as pleasantries (e.g., ``Thank you for that question'' or ``Welcome to the call today'') or references to document sections (e.g., ``We refer to Section \ldots''). To mitigate this, we have developed a detailed preprocessing pipeline to filter out such invalid sentences.

Our process begins with NLTK sentence segmentation \cite{bird2009natural} to extract potential sentence sequences. However, this alone does not guarantee that all extracted sentences meet our criteria for valid premises. We identify and remove potential tables by searching for table-related keywords (e.g., ``table'' or ``Table''), long ellipses (e.g., ``\ldots\ldots''), extended dash sequences (e.g., ``-----''), long underscore sequences (e.g., ``\_\_\_''), and instances where the number of new lines (e.g., \texttt{\textbackslash n}) exceeds 10.

Titles and subtitles are identified by checking if more than 60\% of the words in a sentence start with an uppercase letter. Sentences with over 60\% numerical characters are also removed, along with any sentences containing URLs.

Additionally, we exclude sentences containing specific keywords, such as \texttt{[thank you, thanks, greetings, please, see below, continued, check mark]}. For Annual Reports and SEC filings, we filter out sentences that do not start with a letter or begin with a lowercase letter. Finally, we limit our selection to sentences that are at least 10 words long but no more than 100 words. To ensure a representative sample, we draw from different length bins to capture a range of premise lengths.

\section{Z-Filtering Results}
\label{app:z_filtering}
Table \ref{tab:z_filter} shows the top 10 features with the highest \textit{z-statistic} score before and after applying \textit{Z-filtering} \cite{wu2022generating} on the Llama-generated instances. High lexical correlations with the neutral label are observed, with Llama generating hypotheses containing unigrams such as \textit{influence}, \textit{may}, \textit{might}, and \textit{factors}, reported with high z-scores. After applying the algorithm, lower z-scores are reported across the features, indicating that task-independent features with high label correlations are filtered out. For instance, the unigram \textit{might} is highly correlated with the neutral class (z-statistic = 56.75) but reduces to 25.82 after filtering. However, some correlations remain in the dataset, which models can still exploit, albeit at a reduced level compared to the pre-filtered set. 
\begin{table}[h]
\centering
\small
\scalebox{0.7}{
    \begin{tabular}{lclclc}
      \toprule
          \multicolumn{3}{c}{\bf{Before \textit{Z-filtering}}} & \multicolumn{3}{c}{\bf{After \textit{Z-filtering}}} \\
      \cmidrule(lr){1-3} \cmidrule(lr){4-6}
      {\bf Feature} &  {\bf z-statistics} &  {\bf Label}  & {\bf Feature} &  {\bf z-statistics}  & {\bf Label} \\
      \midrule
        may              & 94.69 & N & overlap > 0.5 & 35.18 & E \\
        may be           & 63.66 & N & overlap > 0.4 & 34.26 & E \\
        influence        & 60.64 & N & overlap > 0.6 & 31.61 & E \\
        influenced       & 60.06 & N & no                & 30.62 & C \\
        influenced by    & 59.87 & N & 'nt               & 27.74 & N \\
        factors          & 58.23 & N & by                & 26.84 & N \\
        flu              & 58.12 & N & might             & 25.82 & C \\
        actors           & 57.90 & N & will be           & 25.80 & N \\
        might            & 56.75 & N & impacted          & 25.77 & N \\
        factor           & 56.25 & N & le                & 25.76 & C \\
      \bottomrule
    \end{tabular}}
\caption{Top 10 biased features with the highest z-statistics before (left) and after (right) applying \textit{Z-filtering} on the Llama-generated instances. The label \textit{N} indicates the neutral class, \textit{C} indicates the contradiction class, and \textit{E} indicates the entailment class.}
\label{tab:z_filter}
\end{table}
Table \ref{tab:z-filt-perf} presents the performance scores of XLM-RoBERTa trained on general-domain NLI datasets before and after applying \textit{z-filtering}. The observed performance drop after \textit{z-filtering} suggests that the removed instances were primarily those the model previously classified correctly, likely due to exploiting biases in the training data. Notably, XLM-RoBERTa has been trained on datasets known to contain spurious correlations \cite{wu2022generating, cosma-etal-2024-hard}. Therefore, the performance drop is expected, as removing these artifacts makes the model less reliant on learned biases.
\begin{table}[H]
\centering
\resizebox{0.8\columnwidth}{!}{%
\begin{tabular}{@{}lllll@{}}
\toprule
 & \multicolumn{2}{l}{\textbf{Before z-fiilter}} & \multicolumn{2}{l}{\textbf{After z-filter}} \\ \midrule
 & \textbf{Acc.} & \textbf{Macro F1} & \textbf{Acc.} & \textbf{Macro F1} \\ \midrule
Calls & 0.83 & 0.83 & 0.79 & 0.77 \\
SEC & 0.84 & 0.84 & 0.79 & 0.75 \\
Reports & 0.85 & 0.85 & 0.82 & 0.79 \\ \bottomrule
\end{tabular}%
}
\caption{Performance scores of XLM-RoBERTa-Large trained on general domain NLI datasets, SNLI and MNLI, before and after applying z-filtering.}
\label{tab:z-filt-perf}
\end{table}

\section{Hyper-parameter Details for Fine-tuning Experiments}
\label{app:params}
We access all models through the HuggingFace library, and conduct training using the PyTorch Lightning framework \footnote{https://lightning.ai/docs/pytorch/stable/}. Each model is trained for a maximum of 20 epochs. Early stopping is employed with a patience of 5 epochs, based on validation accuracy. We apply an exponential learning rate decay, with the decay rate $\gamma = 0.95$. A grid search is performed over the following hyperparameter space: learning rate (\textit{LR}) $ \in \{1e-5, 2e-5, 3e-5, 5e-5\}$ and batch size (\textit{BS}) $ \in \{24, 32, 64\}$. The optimal hyperparameter configuration is selected based on the highest validation accuracy, and this configuration is used to retrain the model across three independent runs, each initialized with different random seeds. The best-performing hyperparameters for each model are provided in Table \ref{tab:hyperparameter_tuning}.
\begin{table}[h]
\centering
\small
\scalebox{1.0}{
    \begin{tabular}{cccl}
     \toprule
         \textbf{Model}&  \textbf{{LR}}& \textbf{BS} &\textbf{Max Length}\\
         \midrule
         \textsc{RoBERTa-Base}&  \(3e-5\) &  64&200\\
         \textsc{FiLM}&  $5e-5$&  64&200\\
         \textsc{RoBERTa-Large}&  $1e-5$&  32&150\\
         \textsc{BART-Large-MNLI}&  $1e-5$&  32&150\\
         \bottomrule
    \end{tabular}}
    \caption{Best-performing hyperparameters for each fine-tuned PLM}
    \label{tab:hyperparameter_tuning}
\end{table}

\label{app:further_error_analysis}
\section{Further Error Analysis}
\subsection{Confusion Matrix}
Table \ref{tab:error-count-extended} shows the frequency of misclassifications across entailment (E), contradiction (C), and neutral (N) classes, with a notable trend of misclassifying neutral as entailment.

\begin{table}[h]
    \centering
    \resizebox{0.7\columnwidth}{!}{%
    \begin{tabular}{@{}lccccccc@{}}
    \toprule
    & \multicolumn{3}{c}{\textbf{GPT-4 test.}} & \multicolumn{3}{c}{\textbf{LLaMA test.}} \\
    \cmidrule(lr){2-4} \cmidrule(lr){5-7}
    \textbf{Labels} & \textbf{E} & \textbf{C} & \textbf{N} & \textbf{E} & \textbf{C} & \textbf{N} \\ \midrule
    \textsc{\bf E} & 880 & 13 & 60 & 373 & 6 & 29 \\
    \textsc{\bf C} & 20 & 539 & 55 & 10 & 289 & 35 \\
    \textsc{\bf N} & \cellcolor{gray!20} 209 & 45 & 236 & \cellcolor{gray!20} 195 & 38 & 272 \\ \bottomrule
    \end{tabular}
    }
    \caption{Confusion matrix for Llama 3.1 70B broken down for GPT-4 and Llama test sets. 
    }
    \label{tab:error-count-extended}
\end{table}
Regardless of which model was used for generation, Llama 3.1 70B often predicts E when the hypothesis expands on the premise without contradiction, making it challenging to determine if the extra information is implied or unrelated.
In such cases, we believe LLMs are inferring more than what is logically warranted. 
Consider example from SEC Fillings,  a premise {\it ``The Corporate General Partner will continue to maintain compliance with franchise agreements and be economically prudent.''}, and its hypothesis {\it ``The Corporate General Partner will prioritize cost-effective decisions to ensure long-term financial sustainability.''}. 
While {\it ``economically prudent''} covers a range of financially responsible behaviours, e.g. {\it ``cost-effective decisions''}. The hypothesis 
indicates that this behaviour is a priority, which is not guaranteed in the premise. 

\subsection{Examples demonstrating challenging reasoning patterns}
\label{app:challenging_instances}
We iteratively refine the prompt exemplars at each generation step by analyzing the predictions of a general-domain NLI model. Table \ref{tab:challenging_examples} presents examples of the different reasoning patterns we have identified. We use these examples as in-context exemplars to guide the LLM in generating instances that necessitate reasoning for accurate inference. Notably, some of the challenging reasoning patterns resemble revisions made by human annotators in the adversarial NLI setting~\cite{nie2019adversarial}. 
\begin{table*}[ht]
    \centering
    \resizebox{0.9\textwidth}{!}{%
    \begin{tabular}{lll}
    \toprule
        \textbf{Example} & \textbf{Label} & \textbf{Reasoning}\\\midrule
        \makecell[l]{
        \textbf{P:} We earned \$535,000, \$606,000, and \$794,000 of management fees under the \\
        agreement for the years ended December 31, 2014, 2013, and 2012. \\
        \textbf{H:} The management fees earned under the agreement in 2014 were less \\
        than those earned in 2013.
        }&\colorbox{con}{\textit{Contradiction}}&\makecell[l]{Mathematical \\ Reasoning}\\\midrule
        \makecell[l]{
        \textbf{P:} Our net equity investment in these VIEs was 808.3 million as of \\
        December 31, 2018, and 606.2 million as of December 31, 2017. \\
        \textbf{H:} At some point during the year 2018, the net equity investment in these \\
        VIEs was either increasing or decreasing from its value at the end of 2017.
        }& \colorbox{neu}{\textit{Neutral}} & \makecell[l]{Temporal \\ Reasoning}\\\midrule
        \makecell[l]{
        \textbf{P:} The Company believes EBITDA is commonly used by financial analysts \\
        and others in the industries in which the Company operates and, thus, \\
        provides useful information to investors. \\
        \textbf{H:} EBITDA is a metric commonly employed by financial industry professionals \\
        to assess a company's financial performance, indicating its significance \\
        in financial analysis.
        }&\colorbox{ent}{\textit{Entailment}}&\makecell[l]{Financial \\ Knowledge}\\\midrule
        \makecell[l]{
        \textbf{P:} Our case growth for Sysco-branded sales to local customers increased \\
        50 basis points in fiscal 2019. \\
        \textbf{H:} There was a modest uptick in our local customer sales last year in 2018.
        }&\colorbox{neu}{\textit{Neutral}}&\makecell[l]{Tricky linguistic \\ construct}\\\midrule
        \makecell[l]{
        \textbf{P:} We've been doubled down in seed for the last four years. \\
        \textbf{H:} We have reduced our seed investment substantially over the past four \\
        years.
        }&\colorbox{con}{\textit{Contradiction}}&\makecell[l]{Idiomatic \\ Expression}\\
    \bottomrule
    \end{tabular}}
    \caption{\textbf{Examples of challenging reasoning patterns} identified through the prompt iteration process and included in the optimised prompts as in-context exemplars to steer LLM generations towards high-quality examples.}
    \label{tab:challenging_examples}
\end{table*}

\section{Relabelled instance examples}
\label{app:relabeled}
Examples of instances relabeled by annotators are presented in Table \ref{tab:relabeled_instances}. We observe that LLMs tend to append additional information not present in the premise sentence to entailment or contradiction-generated hypotheses, resulting in relabeling to neutral. In some cases, the LLM demonstrates a limited understanding of the NLI task and generates an incorrect label for some of the instances.
\begin{table*}[h!]
    \centering
    \resizebox{0.9\textwidth}{!}{%
    \begin{tabular}{lll}
    \toprule
        \textbf{Example} & \textbf{Generated Label} & \textbf{Majority Label}\\\midrule
        \makecell[l]{
        \textbf{P}: If our supply of raw materials is interrupted, our results of \\
        operations could be adversely affected. \\
        \textbf{H}: If our supply of raw materials is interrupted, our  
        results of  \\ operations might be adversely affected.
        } & \colorbox{neu}{\textit{Neutral}} & \makecell[l]{\colorbox{ent}{\textit{Entailment}}} \\\midrule

        \makecell[l]{
        \textbf{P}: For the years ended December 31, 2002, 2003 and 2004, our \\
        net cash provided by (used for) financing activities was \\
        (286.3), 71.1 million and \$(1,686.4) million, respectively. \\
        \textbf{H}: The company's cash flow from financing activities fluctuated \\
        between 2002 and 2004.
        } & \colorbox{neu}{\textit{Neutral}} & \makecell[l]{\colorbox{ent}{\textit{Entailment}}} \\\midrule
        
        \makecell[l]{
        \textbf{P}: A Look Ahead As we enter 2007, we have identified several \\
        priorities by which to grow our business. \\
        \textbf{H}: The company has outlined a strategic plan for business \\
        expansion in 2007, indicating a proactive approach towards \\
        growth and development, \textcolor{red}{which is likely to involve investment} \\
        \textcolor{red}{in new opportunities, resource allocation, and possibly} \\
        \textcolor{red}{restructuring of certain business segments.}
        } & \colorbox{ent}{\textit{Entailment}} & \makecell[l]{\colorbox{neu}{\textit{Neutral}}} \\\midrule

        \makecell[l]{
        \textbf{P}: In November 2016 the LPSC submitted testimony \\ disputing certain aspects of the calculations. \\
        \textbf{H}: The LPSC's submission of testimony in November 2016 was the sole\\ factor that \textcolor{red}{{led to the immediate acceptance of the calculations without}} \\ \textcolor{red}{{any further review or revision.}}
        } & \colorbox{con}{\textit{Contradiction}} & \makecell[l]{\colorbox{neu}{\textit{Neutral}}} \\
    \bottomrule
    \end{tabular}}
    \caption{\textbf{Examples of relabeled instances:} The first two examples show instances where LLMs generate incorrect labels while the last two examples show instances where LLMs tend to include additional information not in the premise sentence  Additional information not in the premise is shown in \textcolor{red}{red}.}
    \label{tab:relabeled_instances}
\end{table*}

\clearpage
\onecolumn
\section{LLM Experiments}
\label{app:prompt}
\subsection{Running Inferences}
All the models are accessed from Hugging Face\footnote{https://huggingface.co/}, and inferences are run using vLLM \cite{kwon2023efficient} if a model is supported. We run inferences on all the models using full precision, except for Llama 3.1 70B, for which we use a 4-bit quantized version to reduce memory footprint. Additionally, we run the FinMA models with the \texttt{Guidance}\footnote{https://github.com/guidance-ai/guidance} library to constrain the output to only the NLI class labels. During the initial runs of the FinMA models \cite{xie2023pixiu}, we observed that the LLMs frequently failed to follow the task instructions and instead generated hallucinated class labels, such as \textit{positive} and \textit{negative}, possibly due to overfitting to sentiment analysis tasks in the instruction-tuned data. We run all inferences using a temperature value of \(0\) across three independent runs with randomly initialized seeds.

\subsection{Prompts}
We provide the prompt templates used in the paper across the generation and evaluation experiments below. 

\begin{prompt}[width=\textwidth, title={Zero-Shot (ZS)}]
You will be given pairs of sentences: a premise and a hypothesis. Assign one of the following labels to each pair: `entailment', `neutral', or `contradiction'.  

Return only the label.

Premise: \texttt{\{premise\}}  

Hypothesis: \texttt{\{hypothesis\}}  

Label:
\end{prompt}

\vspace{10pt}
\begin{prompt}[width=\textwidth, title={Zero-Shot w/ Annotation Guidelines (ZS-AG)}]
You will be given pairs of sentences: a premise and a hypothesis. The premise sentence comes from one of three types of financial texts:

\begin{itemize}
    \setlength{\itemsep}{0pt}
    \item A company’s SEC filings
    \item A company’s annual report
    \item A company’s earnings call transcript
\end{itemize}

Your task is to assign one of the following labels to each pair based on these definitions:  

\begin{itemize}
    \setlength{\itemsep}{0pt}   
    \item \textbf{entailment}: The truth of the premise guarantees the truth of the hypothesis.
    \item \textbf{neutral}: The truth of the premise neither guarantees nor contradicts the truth of the hypothesis.
    \item \textbf{contradiction}: The truth of the premise guarantees that the hypothesis is false.
\end{itemize}

Given the premise and hypothesis below, determine the relationship between the sentences and return `entailment', `neutral', or `contradiction':  

Premise: \texttt{\{premise\}}  

Hypothesis: \texttt{\{hypothesis\}}  

Label:
\end{prompt}

\begin{prompt}[width=\textwidth, title={Few-Shot w/ Annotation Guidelines (Few-Shot-AG)}]
You will be given pairs of sentences: a premise and a hypothesis. The premise sentence comes from one of three types of financial texts:

\begin{itemize}
    \setlength{\itemsep}{0pt}
    \item A company’s SEC filings
    \item A company’s annual report
    \item A company’s earnings call transcript
\end{itemize}

Your task is to assign one of the following labels to each pair based on these definitions:  

\begin{itemize}
    \setlength{\itemsep}{0pt}   
    \item \textbf{entailment}: The truth of the premise guarantees the truth of the hypothesis.
    \item \textbf{neutral}: The truth of the premise neither guarantees nor contradicts the truth of the hypothesis.
    \item \textbf{contradiction}: The truth of the premise guarantees that the hypothesis is false.
\end{itemize}

\textbf{Here are a few examples:  \texttt{\{examples\}}}

\begin{enumerate}
    \item \textbf{Premise}: And then the actual funded balances that we get from various other lines \textbf{Hypothesis}: The funded balances obtained from various lines are expected to increase in the next fiscal quarter. \textbf{Label}: neutral 
    \item \textbf{Premise}: Wet remember, I think about 126 stores that we've already remodeled on that new store prototype. \textbf{Hypothesis}: The company has completed the remodeling of approximately 126 stores using the new store prototype. \textbf{Label}: entailment 
    \item \textbf{Premise}: We maintain allowances for doubtful accounts for estimated losses resulting from the inability of customers to make required payments. \textbf{Hypothesis}: We do not anticipate any losses from customers being unable to make their required payments, so no allowances for doubtful accounts are maintained. \textbf{Label}:contradiction
\end{enumerate}

Given the premise and hypothesis below, determine the relationship between the sentences and return `entailment', `neutral', or `contradiction':  

Premise: \texttt{\{premise\}}  

Hypothesis: \texttt{\{hypothesis\}}  

Label:
\end{prompt}

\clearpage

\begin{prompt}[width=\textwidth, title={Chain-of-Thought (CoT) w/ Rationale}]
You will be given pairs of sentences: a premise and a hypothesis. The premise sentence comes from one of three types of financial texts:

\begin{itemize}
    \setlength{\itemsep}{0pt}
    \item A company’s SEC filings
    \item A company’s annual report
    \item A company’s earnings call transcript
\end{itemize}

Your task is to determine the relationship between the premise and hypothesis and assign one of the following labels:  

\begin{itemize}
    \setlength{\itemsep}{0pt}
    \item \textbf{entailment}: The truth of the premise guarantees the truth of the hypothesis.
    \item \textbf{neutral}: The truth of the premise neither guarantees nor contradicts the truth of the hypothesis.
    \item \textbf{contradiction}: The truth of the premise guarantees that the hypothesis is false.
\end{itemize}

To decide on the correct label, follow these steps:

\begin{enumerate}
    \setlength{\itemsep}{0pt}
    \item \textbf{Understand the Premise}: Carefully read and comprehend the premise to identify the key information it provides.
    \item \textbf{Analyze the Hypothesis}: Read the hypothesis and determine what it claims or implies. Consider if it adds new information, rephrases the premise, or predicts something based on the premise.
    \item \textbf{Compare and Reason}: Compare the key information from the premise and hypothesis. Does the hypothesis logically follow from the premise, contradict it, or is it unrelated?
    \item \textbf{Assign the Label}: Based on your reasoning, choose the appropriate label: `entailment', `neutral', or `contradiction'.
    \item \textbf{Explain Your Reasoning}: Briefly describe the thought process that led to your decision.
\end{enumerate}

Here are a few examples: \texttt{\{examples\}}

Given the premise and hypothesis below, determine the relationship between the sentences and return `entailment', `neutral', or `contradiction':  

Premise: \texttt{\{premise\}}  

Hypothesis: \texttt{\{hypothesis\}}  

Reasoning: [Describe your reasoning here]  

Label: 
\end{prompt}

\clearpage
\begin{prompt}[width=\textwidth, title={Data Generation Prompt}]
You are now a \textbf{\texttt{\{role\}}} and will be provided with a sentence from a \textbf{\texttt{\{financial domain\}}}. Using only this sentence and your general knowledge about the world:

\begin{itemize}
    \setlength{\itemsep}{0pt}
    \item Write one \textbf{entailment} hypothesis where the truth of the premise guarantees the truth of the hypothesis.
    \item Write one \textbf{neutral} hypothesis where the truth of the premise neither guarantees nor contradicts the truth of the hypothesis.
    \item Write one \textbf{contradiction} hypothesis where the truth of the premise guarantees that the hypothesis is false.
\end{itemize}

\textbf{Guidelines for Writing High-Quality Hypotheses:}

\begin{enumerate}
    \item \textbf{Numerical Reasoning:}  
    If the premise contains numbers, write hypotheses that require mathematical knowledge or quantitative reasoning. This means the relationship between the premise and the hypothesis should involve some form of mathematical deduction.  
    \textit{Example:} \texttt{\{examples\}}

    \item \textbf{Temporal Reasoning:}  
    If the premise includes dates, write hypotheses that require an understanding of time. The hypothesis should involve an inference that requires understanding dates and the relationship between different points in time.  
    \textit{Example:} \texttt{\{examples\}}

    \item \textbf{Financial Knowledge:}  
    If the premise includes financial concepts, generate hypotheses that require familiarity with financial reasoning.  
    \textit{Example:} \texttt{\{examples\}}

    \item \textbf{Linguistic Knowledge:}  
    Try to write hypotheses that require linguistic understanding, such as antonyms, hypernyms, metaphors, or syntactic ambiguity.  
    \textit{Example:} \texttt{\{examples\}}
\end{enumerate}

\textbf{Important Writing Considerations:}
\begin{itemize}
    \setlength{\itemsep}{0pt}
    \item Write all hypotheses as clear, declarative sentences. Do not phrase them as questions.
    \item For an \textbf{entailment} label, ensure the hypothesis can be directly and fully verified using only the information provided in the premise. Avoid adding assumptions or speculative language such as "likely," "indicating," or "potential."
    \item When generating a \textbf{contradiction} hypothesis, avoid simply negating the premise. Instead, use financial knowledge and linguistic reasoning to construct a meaningful contradiction.
\end{itemize}

\textbf{Given the premise:} \texttt{\{premise sentence\}}  

\textbf{Write three hypotheses using:} \texttt{\{writing style\}}  

\textbf{Return your response in the format below:}  

\begin{itemize}
    \item \textbf{Entailment:} \texttt{\{entailment hypothesis\}}
    \item \textbf{Neutral:} \texttt{\{neutral hypothesis\}}
    \item \textbf{Contradiction:} \texttt{\{contradiction hypothesis\}}
\end{itemize}
\end{prompt}

\section{Annotation Instructions for the Financial Natural Language Inference Dataset (FinNLI)}
\label{app:annotation}
\begin{tcolorbox}[colback=gray!10, colframe=warmnavyblue, title=Annotation Instructions for FinNLI, breakable]
\textbf{Task Overview:}  
You will annotate a dataset for \textbf{Financial Natural Language Inference (FinNLI)}.  

You will be given pairs of sentences: a \textbf{premise} and a \textbf{hypothesis}, where:
\begin{itemize}
    \item The \textbf{premise} comes from financial texts (SEC filings, annual reports, earnings call transcripts).
    \item The \textbf{hypothesis} is generated using a Large Language Model.
\end{itemize}

\textbf{Your task is to:}
\begin{enumerate}
    \item \textbf{Verify that both the premise and hypothesis are valid and meaningful sentences} e.g. are not phrase fragments or nonsense, which would prevent us being able to assign a label to the relationship between the premise and the hypothesis. We have taken steps to preprocess the sampled sentences, so we expect invalid sentences to be a low frequency event. If either of the sentences is invalid, you do not need to provide relationship label for them.
    \item \textbf{Assign one label to each of hypothesis-premise pair.}: Choose one of the three:
    \begin{itemize}
        \item \textbf{Entailment} – A relationship between two statements where the truth of the premise guarantees the truth the hypothesis. The hypothesis follows logically from the information contained in the premise.
        \item \textbf{Neutral} – A relationship between two statements where the truth of the premise neither guarantees nor contradicts the truth of the hypothesis. It is not possible to determine whether the hypothesis is true or false without further information.
        \item \textbf{Contradiction} – The truth of the premise guarantees that the hypothesis is false. The hypothesis is logically false from the information contained in the premise.
    \end{itemize}
    \item \textbf{Indicate your level of confidence in the annotation: Low or High.} Depends on how confident an annotator is while defining the relation between premise and hypothesis.
    \item \textbf{[Optional] Provide Comments} pertaining to annotating the instance, e.g. provide the rationale for the decision in free text form explaining why the instance is difficult, and/or why the confidence is low. Raise any other concerns you have in this field, e.g. in unlikely cases where you consider the texts harmful or biased.
\end{enumerate}

\textbf{Label Definitions and Examples:}

\textbf{Entailment:} The hypothesis logically follows from the premise.
\begin{quote}
\textbf{Premise:} We may acquire hotels in joint ventures with third parties that could result in conflicts. \\
\textbf{Hypothesis:} The company is considering partnerships with third parties for hotel acquisitions. \\
\textbf{Label:} Entailment
\end{quote}

\textbf{Neutral:} The premise neither guarantees nor contradicts the hypothesis.
\begin{quote}
\textbf{Premise:} Director stock options granted under the plans expire after seven years and vest fully after six months. \\
\textbf{Hypothesis:} All employee stock options granted under the plans expire after seven years and vest fully after six months. \\
\textbf{Label:} Neutral
\end{quote}

\textbf{Contradiction:} The truth of the premise guarantees that the hypothesis is false.
\begin{quote}
\textbf{Premise:} The user assumes all risks for any damages or losses arising from any use of this information, except to the extent such damages or losses cannot be limited or excluded by applicable law. \\
\textbf{Hypothesis:} The provider is liable for any damages or losses that the user may incur from using this information. \\
\textbf{Label:} Contradiction
\end{quote}

\textbf{Handling Invalid Sentences:}
If a sentence is not meaningful, classify it as \textbf{Premise\_notValid} or \textbf{Hypothesis\_notValid}. Common cases include:

\begin{enumerate}
    \item \textbf{Incomplete Sentence:} Sentence is visibly incomplete.  
    \textit{Example:} ``Premise: Before joining FMC Technologies, Mr.''
    
    \item \textbf{Complex Notation:} Excessive numbers or financial notation making it unreadable.  
    \textit{Example:} ``Premise: \$700.0 \$613.7 \$492.1 95 Rest of World 108.6 96.0 93.9''
    
    \item \textbf{Document Title or Subtitle:} Appears to be a section title instead of a standalone sentence.  
    \textit{Example:} ``Net Cruise Cost less fuel expense.''
    
    \item \textbf{Reference to External Text:} Refers to another document section without self-contained meaning.  
    \textit{Example:} ``For further discussion of our annual impairment test, see the Critical Accounting Policies and Estimates section.''
\end{enumerate}

\textbf{Edge Cases:}
\begin{itemize}
    \item \textbf{Labels Are Not Symmetrical:}
    \begin{quote}
        \textbf{Premise:} Girl ate an apple. \\
        \textbf{Hypothesis:} Girl ate a fruit. \\
        \textbf{Label:} Entailment \\

        \textbf{Premise:} Girl ate a fruit. \\
        \textbf{Hypothesis:} Girl ate an apple. \\
        \textbf{Label:} Neutral
    \end{quote}
    
    \item \textbf{Use of Domain Knowledge:} Some financial knowledge is allowed.
    \begin{quote}
        \textbf{Premise:} The Company believes EBITDA is commonly used by financial analysts and provides useful information to investors. \\
        \textbf{Hypothesis:} EBITDA is a metric commonly used by financial professionals to assess financial performance. \\
        \textbf{Label:} Entailment
    \end{quote}
    
    \item \textbf{Handling Questions:}  
    \begin{quote}
        \textbf{Premise:} Jane is coming at 6. \\
        \textbf{Hypothesis:} When is Jane coming? \\
        \textbf{Label:} Entailment

        \textbf{Premise:} Jane is coming at 6. \\
        \textbf{Hypothesis:} Why isn’t Jane coming? \\
        \textbf{Label:} Contradiction
    \end{quote}
\end{itemize}

\textbf{Final Notes:}
\begin{itemize}
    \item If uncertain, mark annotation confidence as \textbf{Low}.
    \item Avoid unnecessary speculation beyond the provided premise.
    \item If the annotation process reveals bias or harmful content, leave a comment.
\end{itemize}
\end{tcolorbox}

\end{document}